%% file: DovSG.tex
\documentclass[letterpaper, 10 pt, journal, twoside]{IEEEtran}

\IEEEoverridecommandlockouts                              

\usepackage{amsmath} 
\usepackage{amssymb}  
\usepackage{bm}
\usepackage{multirow}
\usepackage{makecell}
\usepackage{bbding}

\usepackage{graphics} 
\usepackage{epsfig} 
\let\labelindent\relax

\usepackage{mathtools}
\usepackage{enumitem}
\usepackage{times} 

\usepackage[utf8]{inputenc} 
\usepackage[T1]{fontenc}    
\usepackage{hyperref}       
            
\usepackage{url}            
\usepackage{booktabs}       
\usepackage{amsfonts}       
\usepackage{nicefrac}       
\usepackage{microtype}      
\usepackage{xcolor}         
\usepackage{cleveref}
\usepackage{hyperref}
\usepackage{graphicx}
\usepackage{float}
\usepackage{multirow}
\usepackage{diagbox}

\usepackage{algorithm}
\usepackage{algpseudocode}
\usepackage{arydshln}
\usepackage{marvosym}

\usepackage{multirow}
\usepackage{listings}
\usepackage{graphicx}
\usepackage{siunitx} 
\usepackage{capt-of}
\usepackage{cuted}

\usepackage{xcolor}

\begin{document}

\markboth{IEEE Robotics and Automation Letters. Preprint Version. February, 2025}
{Yan \MakeLowercase{\textit{et al.}}: Dynamic Open-Vocabulary 3D Scene Graphs for Long-term Language-Guided Mobile Manipulation}


\title{Dynamic Open-Vocabulary 3D Scene Graphs for Long-term Language-Guided Mobile Manipulation}


\author{Zhijie Yan$^{1}$, Shufei Li$^{2}$, Zuoxu Wang$^{1}$, Lixiu Wu$^{3}$, Han Wang$^{4}$, Jun Zhu$^{4}$, Lijiang Chen$^{4}$, Jihong Liu$^{1}$%
\thanks{Manuscript received: October 16, 2024; Revised January 2, 2025; Accepted February 4, 2025.}%
\thanks{This paper was recommended for publication by Editor Gentiane Venture upon evaluation of the Associate Editor and Reviewers’ comments.}%
\thanks{This work was supported by the National Natural Science Foundation of China (NSFC, Grant No. 52205244), the Ministry of Industry and Information Technology (MIIT) Key Laboratory of Intelligent Manufacturing for High-end Aerospace Products, and the Beijing Key Laboratory of Digital Design and Manufacturing.}%
\thanks{Corresponding author: Zuoxu Wang, e-mail: zuoxu\_wang@buaa.edu.cn.}%
\thanks{$^{1}$Zhijie Yan, Zuoxu Wang and Jihong Liu are with School of Mechanical Engineering and Automation, Beihang University, Beijing 100191, China.}%
\thanks{$^{2}$Shufei Li is with Department of Systems Engineering, City University of Hong Kong, Hong Kong SAR 518057, China.}%
\thanks{$^{3}$Lixiu Wu is with School of Information Engineering, Minzu University of China, Beijing 100081, China.}%
\thanks{$^{4}$Han Wang, Jun Zhu and Lijiang Chen are with Afanti Tech LLC, Beijing 100192, China.}%
\thanks{Digital Object Identifier (DOI): see top of this page.}
}


\maketitle

\newcommand{\ourname}{\texttt{DovSG}}

\begin{strip}
    \centering
    \vspace{-51mm}
    \includegraphics[width=\linewidth]{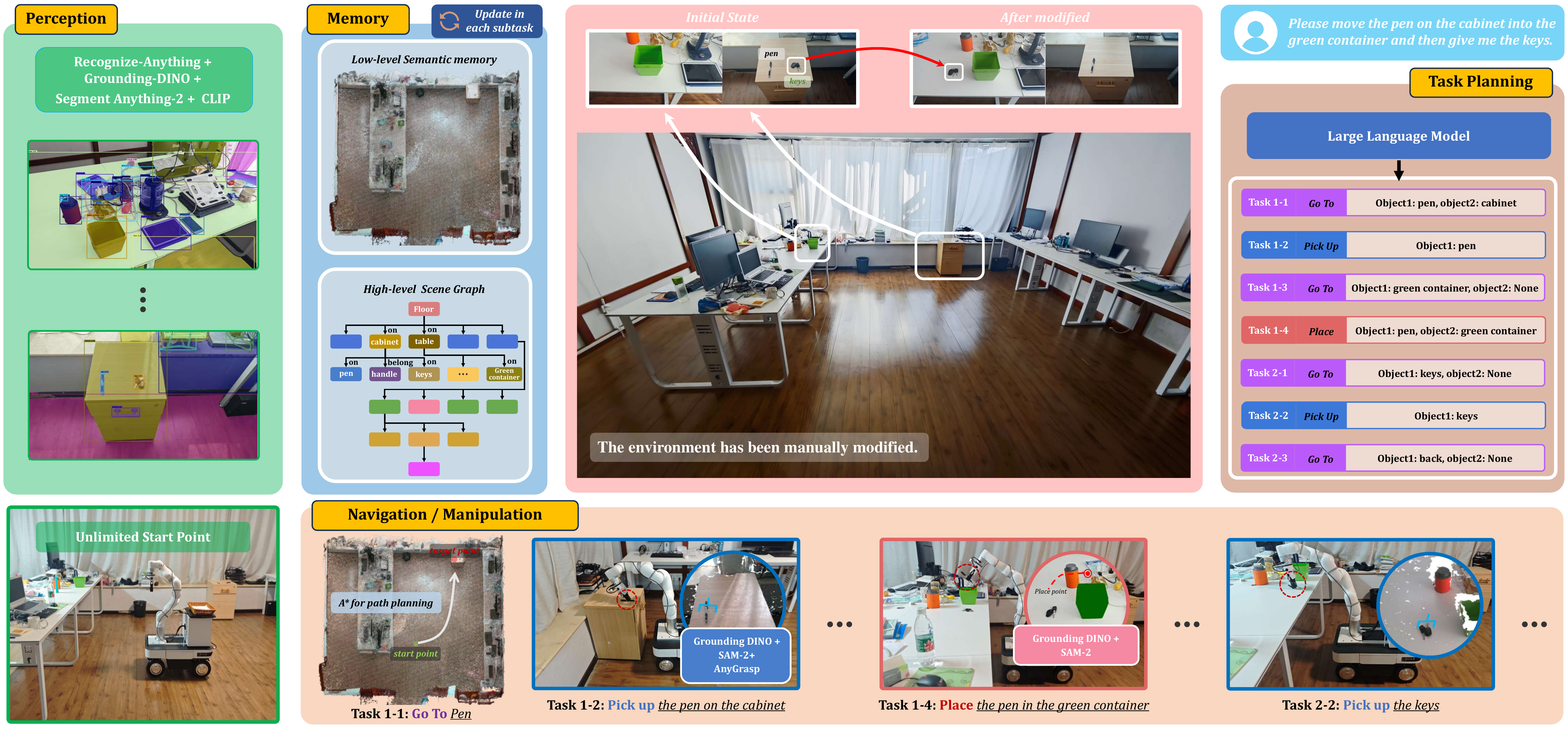}
    \vspace{-7mm}
    \captionof{figure}{\textbf{Overview of Our \ourname{} System.} \ourname{} is a mobile robotic system designed to perform long-term tasks in real-world environments. It can detect changes in the scene during task execution, ensuring that subsequent subtasks are completed correctly. The system consists of five main components: perception, memory, task planning, navigation, and manipulation. The memory module includes a lower-level semantic memory and a higher-level scene graph, both of which are continuously updated as the robot explores the environment. This enables the robot to promptly detect manual changes (e.g., keys being moved from cabinet to table) and make the necessary adjustments for subsequent tasks (such as correctly executing Task 2-2).
    }
    \label{fig:teaser}
    \vspace{-3mm}
\end{strip}


\input{a_abstract}
\input{b_introduction}
\input{c_related_works}
\input{d_method}

\input{e_experiments}
\input{f_limitations}

\input{g_conclusions}

\input{DovSG.bbl}

\end{document}

%% file: a_abstract.tex
\begin{abstract}
Enabling mobile robots to perform long-term tasks in dynamic real-world environments is a formidable challenge, especially when the environment changes frequently due to human-robot interactions or the robot's own actions. Traditional methods typically assume static scenes, which limits their applicability in the continuously changing real world.
To overcome these limitations, we present \ourname{}, a novel mobile manipulation framework that leverages dynamic open-vocabulary 3D scene graphs and a language-guided task planning module for long-term task execution. 
\ourname{} takes RGB-D sequences as input and utilizes vision-language models (VLMs) for object detection to obtain high-level object semantic features. Based on the segmented objects, a structured 3D scene graph is generated for low-level spatial relationships. Furthermore, an efficient mechanism for locally updating the scene graph, allows the robot to adjust parts of the graph dynamically during interactions without the need for full scene reconstruction.  
This mechanism is particularly valuable in dynamic environments, enabling the robot to continually adapt to scene changes and effectively support the execution of long-term tasks. 
We validated our system in real-world environments with varying degrees of manual modifications, demonstrating its effectiveness and superior performance in long-term tasks.
Our project page is available at: \textcolor{blue!80}{\href{https://bjhyzj.github.io/dovsg-web}{https://bjhyzj.github.io/dovsg-web}}.
\end{abstract}


%% file: b_introduction.tex
\section{Introduction}
\IEEEPARstart{M}{obile} manipulation in real-world environments is increasingly expected to perform complex, long-term tasks that require adaptability and resilience to changes. These dynamic environments are characterized by frequent alterations caused by human activities, robot interactions, or the inherent variability of the surroundings. Traditional robotic systems often fall short in such settings because they typically rely on the assumption of a static or minimally changing environment \cite{rt2, driess2023palme, Liu_2024, gilles2022gohome, WANG2025100759}. This limitation restricts their applicability in real-world scenarios where adaptability is crucial.

In this work, we enhance robotic capabilities by introducing a novel and practical robotic framework, the \ourname{} system. This framework comprises five key modules: perception, memory, task planning, navigation, and manipulation, as illustrated in Fig.~\ref{fig:teaser}.
To address the challenge of scene perception, our \textbf{perception module} integrates advanced tools such as Recognize-Anything \cite{zhang2023recognize}, Grounding DINO \cite{liu2024groundingdinomarryingdino}, Segment Anything-2 \cite{ravi2024sam2segmentimages}, and CLIP \cite{radford2021learningtransferablevisualmodels} to detect objects or components and extract their semantic features. In the \textbf{memory module}, each object within the scene is represented as a node characterized by geometric and semantic features, with the relationships between objects encoded in the graph’s edges. 
The module continuously updates object features and scene graphs by locally refining the areas where the robot interacts, preserving knowledge to support ongoing exploration and exploitation, while avoiding the need to reconstruct the entire scene and enabling more efficient adaptation to dynamic environments.
Our \textbf{task planning module} uses the advanced large language models to decompose tasks into manageable subtasks. Then, the \textbf{navigation} and \textbf{manipulation module} are activated to execute the planned actions. 
In experiments, we deployed a mobile robot platform in multiple real-world indoor environments, conducting long-term task trials under varying degrees of manual environmental modifications. We demonstrated that \ourname{} can accurately update the 3D scene graph and perform excellently in long-term tasks and subtasks such as pick-up, place, and navigation. Our \textbf{contributions} are as follows: 

\begin{itemize} 
\item We propose a novel robotic framework that integrates dynamic open-vocabulary 3D scene graphs with language-guided task planning, enabling accurate long-term task execution in dynamic and interactive environments. 
\item We construct dynamic 3D scene graphs that capture rich object semantics and spatial relations, performing localized updates as the robot interacts with its environment, allowing it to adapt efficiently to incremental modifications.
\item We develop a task planning method that decomposes complex tasks into manageable subtasks, including pick-up, place, and navigation, enhancing the robot’s flexibility and scalability in long-term missions. 
\item We implement \ourname{} on real-world mobile robots and demonstrate its capabilities across dynamic environments, showing excellent performance in both long-term tasks and subtasks like navigation and manipulation. \end{itemize}

%% file: c_related_works.tex
\section{Related Works}

\subsubsection{3D Scene Representations for Perception}
3D scene representation in robotics often follows two main approaches: using foundation models to create 3D structures \cite{shen2023distilledfeaturefieldsenable, kerr2023lerflanguageembeddedradiance, zheng2024gaussiangrasper3dlanguagegaussian}, or combining 2D image and vision-language models \cite{liu2024groundingdinomarryingdino, radford2021learningtransferablevisualmodels, ravi2024sam2segmentimages, zhang2023recognize, yan2024manufvissgg} to connect with the 3D world,  showing impressive results in open-vocabulary tasks \cite{gu2023conceptgraphsopenvocabulary3dscene, jiang2024roboexpactionconditionedscenegraph},
enabling language-guided object grounding \cite{zheng2024gaussiangrasper3dlanguagegaussian, Liu_2024, liu2024dynamemonlinedynamicspatiosemantic, gu2023conceptgraphsopenvocabulary3dscene} and 3D reasoning \cite{peng2023openscene3dsceneunderstanding, li2021toward, LI2023102510}. However, dense semantic features for each point are redundant, memory-intensive, and not easily decomposable, limiting their use in dynamic robotics applications.

\subsubsection{3D scene graphs for memory}
3D scene graphs use a hierarchical graph structure to represent objects as nodes and their relationships as edges within a scene \cite{wu2020localize}. This approach efficiently stores semantic information in robot memory, even in large and dynamic environments. Recent methods like ConceptGraphs \cite{gu2023conceptgraphsopenvocabulary3dscene} and HOV-SG \cite{werby23hovsg} create more compact and efficient scene graphs by merging features of the same object across multiple views, thereby reducing memory redundancy. However, these methods often assume static environments and overlook scene updates during interactive robot manipulations. RoboEXP \cite{jiang2024roboexpactionconditionedscenegraph} addresses this by encoding both spatial relationships and logical associations that reflect the effects of robot actions, enabling the discovery of objects through interaction and supporting dynamic scene updates. Building on this, we have applied these concepts to mobile robots, allowing them to rapidly update memory and adjust the 3D scene graph in real-time during dynamic interactions with environments.

\subsubsection{Large Language Models for Planning}
Large language models (LLMs) \cite{openai2023gpt4, geminiteam2023gemini} and VLMs demonstrate significant potential for zero-shot planning in robotics \cite{Yan_2023_ICCV, zheng2024largelanguagemodelspowered}. These models have been used to generate trajectories and plan manipulations, improving robot adaptability. Integrating LLMs with 3D scene graphs offers further opportunities for task planning. In \ourname{}, we leverage GPT-4 to decompose tasks into subtasks that can be executed through graph memory, enabling robots to handle complex tasks flexibly and adaptively.

\subsubsection{Indoor Visual Localization for Interaction}
Robot interaction relies on accurate tracking of 6-DoF poses using maps constructed from image sequences. Visual relocalization techniques can be categorized into structure-based and learning-based methods. Structure-based approaches employ local descriptors to match 2D pixels to 3D scene coordinates and use PnP algorithms for pose recovery, supported by image retrieval and advanced matching techniques like LightGlue \cite{lindenberger2023lightglue} or MatchFormer \cite{wang2022matchformerinterleavingattentiontransformers}. While effective for large-scale environments, they struggle in small, static indoor settings due to expanding image and feature databases. Learning-based methods, such as ACE \cite{brachmann2023acceleratedcoordinateencodinglearning} and DSAC* \cite{9394752}, predict poses via direct regression, enabling rapid end-to-end optimization but with limited precision due to reliance on approximate pose estimates. These methods also require scene-specific training, which limits scalability. 
To achieve fast and accurate relocalization in indoor environments, we combine the strengths of both approaches. 
We use ACE to estimate an initial pose from 2D images, employ LightGlue to match the most similar historical 2D images and their poses, and then further refine the pose by iterative closest point (ICP \cite{besl1992method}).

%% file: d_method.tex
\newcommand{\imgsequence}{\mathcal{I}}
\newcommand{\scenegraph}{\mathcal{G}}
\newcommand{\obj}{\mathbf{o}}
\newcommand{\objset}{\mathbf{O}}
\newcommand{\edg}{\mathbf{e}}
\newcommand{\edgeset}{\mathbf{E}}

\newcommand{\Transform}{T}
\newcommand{\img}{I}

\newcommand{\icls}{\mathbf{c}}
\newcommand{\ibbox}{\mathbf{b}}
\newcommand{\imask}{\mathbf{m}}

\newcommand{\feat}{f}
\newcommand{\visfeat}{\feat^{\text{rgb}}}
\newcommand{\textfeat}{\feat^{\text{text}}}

\newcommand{\clipembed}{\text{Embed}}
\newcommand{\pcd}{\textit{pcd}}

\newcommand{\poseset}{\mathcal{P}}

\newcommand{\irgb}{\img^{\text{rgb}}}
\newcommand{\idep}{\img^{\text{depth}}}
\newcommand{\ipos}{\img^{\text{pose}}}
\newcommand{\icb}{\img^{\text{c2b}}}

\newcommand{\sminus}{\text{-}}

\newcommand{\similarity}{s}
\newcommand{\geoSim}{\mathbf{\similarity_\text{geo}}}
\newcommand{\visSim}{\mathbf{\similarity_\text{vis}}}
\newcommand{\textSim}{\mathbf{\similarity_\text{text}}}
\newcommand{\nnrate}{\text{Nnrate}}

\newcommand{\weight}{\mathbf{w}}

\newcommand{\numDet}{n}

\newcommand{\area}{\text{Area}}
\newcommand{\alphashape}{\alpha\text{-shape}}

\section{Method}

\ourname{} enables mobile robots to perform long-term tasks in indoor environments by constructing dynamic 3D scene graphs and using large language models for task planning. The process starts with scanning the environment using an RGB-D camera to capture images, followed by open-vocabulary 3D object mapping that detects, associates, and fuses objects into a 3D representation. From these, a 3D scene graph is generated, capturing object relationships and continuously updated when the environment changes. Task planning is performed through language-guided decomposition of long-term tasks into subtasks, which are executed via navigation and manipulation modules.

\subsection{Home Scanning and Coordinate Transformation}


\subsubsection{Home Scanning} Following \cite{gilles2022gohome}, we captured a video of the room using an Intel Realsense D455 camera, resulting in a sequence of RGB-D images, $\imgsequence_t = { \img_1, \img_2, \ldots, \img_t }$, $\img_t = \langle \irgb_t, \idep_t \rangle$ (color and depth), with $t$ as the sequence length. The capture focused on both objects of interest and navigable surfaces, especially around objects and containers.

\subsubsection{Coordinate Transformation}
\begin{figure*}
    \centering
    \includegraphics[width=\linewidth]{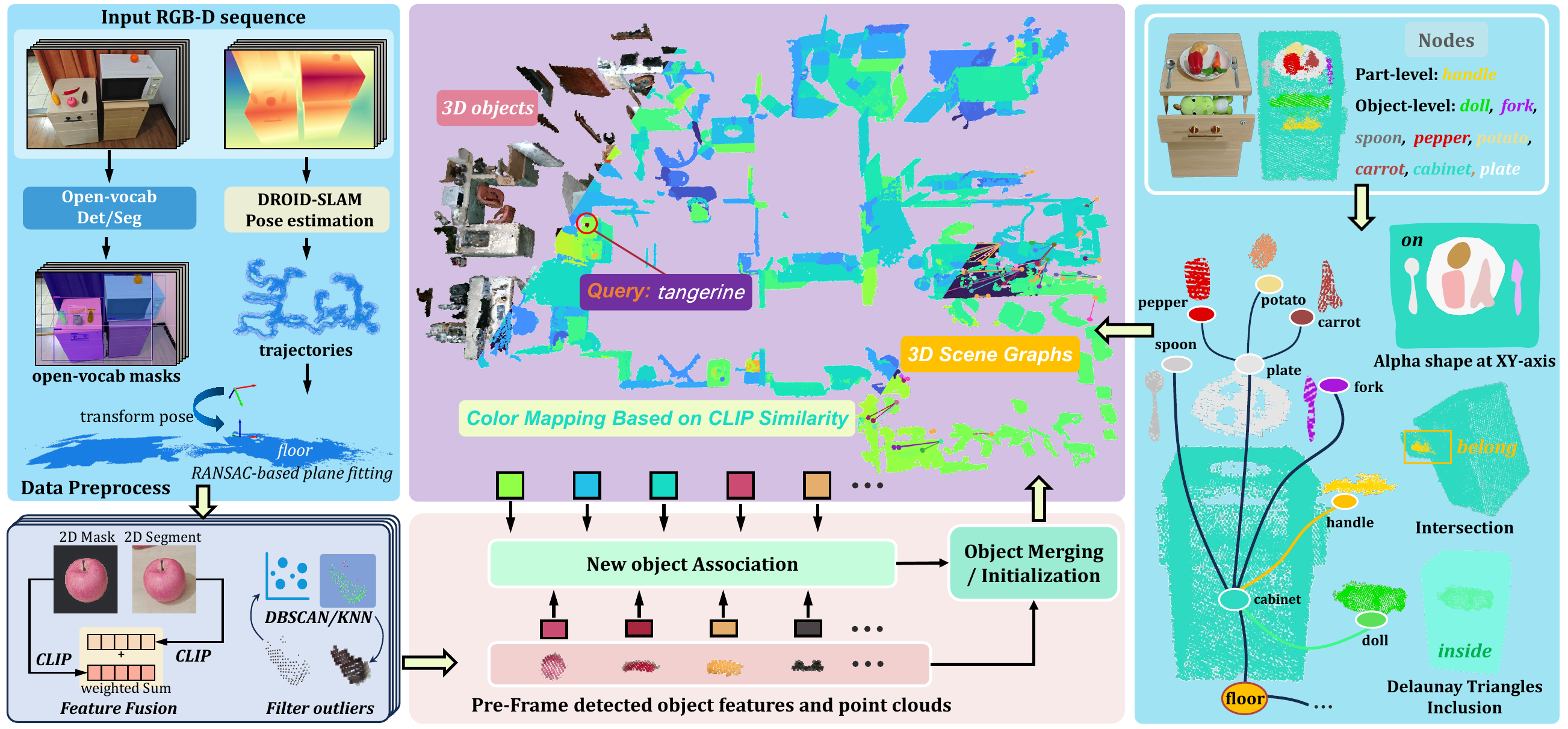}
    \vspace{-7mm}
    \caption{\textbf{Initialization and Construction of 3D Scene Graphs.}
    We first use the RGB-D-based DROID-SLAM \cite{teed2021droid} model to predict the pose of each frame in the scene. Then, we apply an advanced Open-Vocal segmentation model to segment regions in the RGB images, extract semantic feature vectors for each region, and project them onto a 3D point cloud. Based on semantic, geometric, and CLIP feature similarities, the same object captured from multiple views is gradually associated and fused, resulting in a series of 3D objects. Next, we infer the relationships between objects based on their spatial positions and generate edges connecting these objects, forming a scene graph. This scene graph provides a structured and comprehensive understanding of the scene, allowing efficient localization of target objects and enabling easy reconstruction and updating in dynamic environments, and supports task planning for large language models.
    }
    \label{fig:scenegraphinitial}
    \vspace{-5mm}
\end{figure*}

After data collection, we used DROID-SLAM to estimate the camera poses throughout the sequence, as shown in Fig.~\ref{fig:scenegraphinitial} (top left). To achieve accurate pose estimation at true scale, we replaced DROID-SLAM's depth prediction with actual depth data from the camera sensor, resulting in camera poses, $\poseset^\text{droid}$.
In many SLAM algorithms \cite{teed2021droid, murORB2, Zhu2022CVPR}, the first frame's pose is used as the origin, making subsequent poses relative to it, which is often unsuitable for spatial relationships. To normalize the scene, we used the `floor' as a reference and processed each RGB frame with Grounding DINO and Segment Anything-2 to obtain a floor mask. Since the query only focused on the floor, this process was computationally efficient. Once we obtained the mask, we projected 2D pixels to a 3D point cloud.
To align the scene relative to the detected floor, we applied RANSAC to fit a plane to the 3D floor points and computed a transformation matrix $\Transform^\text{floor}$, aligning the floor with the global $z = 0$ plane. Each pose in $\poseset^\text{droid}$ was then transformed as: $\poseset = \mathcal{R}_x \Transform^\text{floor} \poseset^\text{droid}$, where $\mathcal{R}_x$ aligns the coordinate system to the x-axis. This yielded the final set of poses, $\poseset = \{\ipos_1, \ipos_2, \ldots, \ipos_t\}$, with the floor properly aligned. By integrating $\poseset$ with the image sequence $\imgsequence_t$, we obtained $\img_t = \langle \irgb_t, \idep_t, \ipos_t \rangle$.

\subsection{Open-vocabulary 3D Object Mapping}
\label{3d_mapping}
With the RGB-D sequences and camera poses, we proceed to construct an object-centric 3D representation from RGB-D observations $\imgsequence_t = \{ \img_1, \img_2, \ldots, \img_t \}$ to objects $\objset_t=\{\obj_j\}$, $j \in \{1,...,J\}$ is all object length, each object $\obj_{j}$ is characterized by a 3D point cloud $\pcd_{\obj_{j}}$, visual feature $\visfeat_{\obj_j}$ and text feature $\textfeat_{\obj_j}$.
This map is built incrementally, incorporating each incoming frame 
$\img_t$ into the existing object set $\objset_{t-1}$, 
by either adding to existing objects or instantiating new ones.

\subsubsection{Open-vocabuary 2D Segmentation}
\label{ov2dseg}
To maximize object recognition in the scene, we first apply the image tagging model Recognize-Anything \cite{zhang2023recognize} to each frame $\img_t$, generating a set of object classes $\{\icls_{t,i}\}$, $i \in \{1,...,M\}$ detected in the image. We then use $\{\icls_{t,i}$\}, $i \in \{1,...,M\}$  as input to the 2D detector Grounding DINO to obtain object bounding boxes $\{\ibbox_{t,i}\}$, $i \in \{1,...,M\}$ . Finally, we refine these bounding boxes into object masks $\{\imask_{t,i}\}$, $i \in \{1,...,M\}$  using the advanced segmentation model Segment Anything-2. For each obtained 2D mask $\imask_{t,i}$, we extract a cropped image based on its bounding box, as well as an isolated mask image without background, as illustrated in Fig.~\ref{fig:scenegraphinitial} (bottom-left). We then extract the visual features of each object using two mask-based images with CLIP, and fuse them using a weighted sum method, as described in HOV-SG \cite{werby23hovsg}. This combines the CLIP features from both the cropped target image and the isolated mask image to generate a visual feature descriptor:
\begin{equation}
\visfeat_{t,i} = \clipembed(\irgb_t, \ibbox_{t,i}, \imask_{t,i}),
\end{equation}
while the text descriptor is obtained by:
\begin{equation}
\textfeat_{t,i} = \clipembed(\icls_{i,t}),
\end{equation}
each masked region is then projected into 3D, denoised using DBSCAN clustering with an adaptively computed \(\varepsilon\) parameter based on the sorted distances to the \(k\)-nearest neighbors, and transformed to the map frame, resulting in a point cloud $\pcd_{t,i}$, along with unit-normalized semantic feature $\visfeat_{t,i}$ and $\textfeat_{t,i}$.

\subsubsection{Object Association} For every newly detected object $\langle \pcd_{t,i}, \visfeat_{t,i}, \textfeat_{t,i} \rangle$, we compute geometric and semantic similarity with respect to all objects $\obj_{t\sminus1,j}=\langle \pcd_{\obj_j}, \visfeat_{\obj_j}, \textfeat_{\obj_j}, \rangle$ in the map that shares any partial geometric overlap.
The geometric similarity: 
\begin{equation}
\geoSim(i, j)=\nnrate (\pcd_{t,i} , \pcd_{\obj_j} )
\end{equation}
is defined as the ratio of the number of points in point cloud $\pcd_{t,i}$ that have nearest neighbors in point cloud $\pcd_{\obj_j}$, within a distance threshold of $\delta_{\text{nn}}$, to the total number of points in $\pcd_{t,i}$.
The visual and text similarity:
\begin{equation}
\visSim(i, j)=(\visfeat_{t,i})^\top \visfeat_{\obj_j} /2+1/2
\end{equation}
\begin{equation}
\textSim(i, j)=(\textfeat_{t,i})^{\top} \textfeat_{\obj_j} /2+1/2
\end{equation}
is the normalized cosine distance between the corresponding visual descriptors.
The overall similarity measure $\similarity(i, j)$ is a weighted sum of the individual similarity measures: 
\begin{equation}
\similarity(i, j) = \visSim(i, j) \cdot \omega_{\text{v}} + \geoSim(i, j) \cdot \omega_{\text{g}} + \textSim(i, j) \cdot \omega_{\text{t}},
\end{equation}
where $\omega_{\text{v}} + \omega_{\text{g}} + \omega_{\text{t}} = 1$. 
We perform object association by a greedy assignment strategy where each new detection is matched with an existing object with the highest similarity score.
If no match is found with a similarity higher than $\delta_{\text{sim}}$, we initialize a new object.

\subsubsection{Object Fusion} If a detection $\obj_{t\sminus1,j}$ is associated with a mapped object $\obj_j$, we fuse the detection with the map. 
This is achieved by updating the object's features as
\begin{equation}
\visfeat_{\obj_j} = (\numDet_{\obj_j} \visfeat_{\obj_j} + \visfeat_{t,i}) / (\numDet_{\obj_j}+1),
\end{equation}
where $\numDet_{\obj_j}$ is the number of detections that have been associated to $\obj_j$ so far;
and updating the pointcloud as $\pcd_{t,i} \cup \pcd_{\obj_j}$, followed by down-sampling to remove redundant points.

\subsection{3D Scene Graph Generation}
\label{3d_sgg}
\ourname{} constructs a 3D scene graph $\scenegraph_t = \langle \objset_t, \edgeset_t \rangle$, where and $\edgeset_t = \{\edg_{k}\}$, $k \in \{1,...,K\}$ represent the sets of objects and edges, respectively. Given the set of 3D objects $\objset_T$ obtained from the previous step, we estimate their spatial relationships, i.e., the edges $\edgeset_T$, to complete the 3D scene graph, as shown in Fig.~\ref{fig:scenegraphinitial} (column 3). Leveraging the transformation of the scene's coordinate system from the previous steps—where the ground plane serves as the origin and the z-axis points upwards—we can efficiently extract the fundamental spatial relationships among the objects.
While focusing on these relationships, we first voxelize the point cloud $\pcd_{\obj_j}$ of each object $\obj_j$ in the scene. This reduces computation and storage by converting dense point clouds into sparse voxel grids, ensuring efficient scene updates in future tasks. 
We focus on three relationships: ``on'', ``belong'', and ``inside'', as mentioned in RoboEXP \cite{jiang2024roboexpactionconditionedscenegraph}, as these relationships are sufficient for downstream tasks. The ``on'' relationship denotes a stacking or positional hierarchy between objects, such as an apple being placed on a table. The ``belong'' relationship captures ownership or attachment between objects, like a refrigerator handle belonging to the refrigerator. Lastly, the ``inside'' relationship is particularly relevant during the robot's exploration when it opens containers such as drawers or cabinets, revealing that certain objects are contained within others. It is important to note that for ``inside'' relationship, we limit our focus to small-scale containment between objects, as technically, every object in the scene could be considered ``inside'' the overall home environment.

\subsection{Dynamic Scene Adaptation}
In dynamic indoor interaction scenarios, the layout of the environment changes due to human activities or the robot's task execution. These changes are often invisible to previous approaches \cite{Liu_2024, werby23hovsg, gu2023conceptgraphsopenvocabulary3dscene}, and in such environments, if the robot cannot dynamically update its memory, it will soon face failure. To address this issue, we have designed a simple memory update module that can quickly perform local updates to the memory based on new RGB-D observations captured by the robot, as shown in Fig.~\ref{fig:reloc_and_updates}.

\subsubsection{Relocalization and refinement}
\label{reloclizetion_and_refinement}
\begin{figure}
    \centering
    \includegraphics[width=1\linewidth]{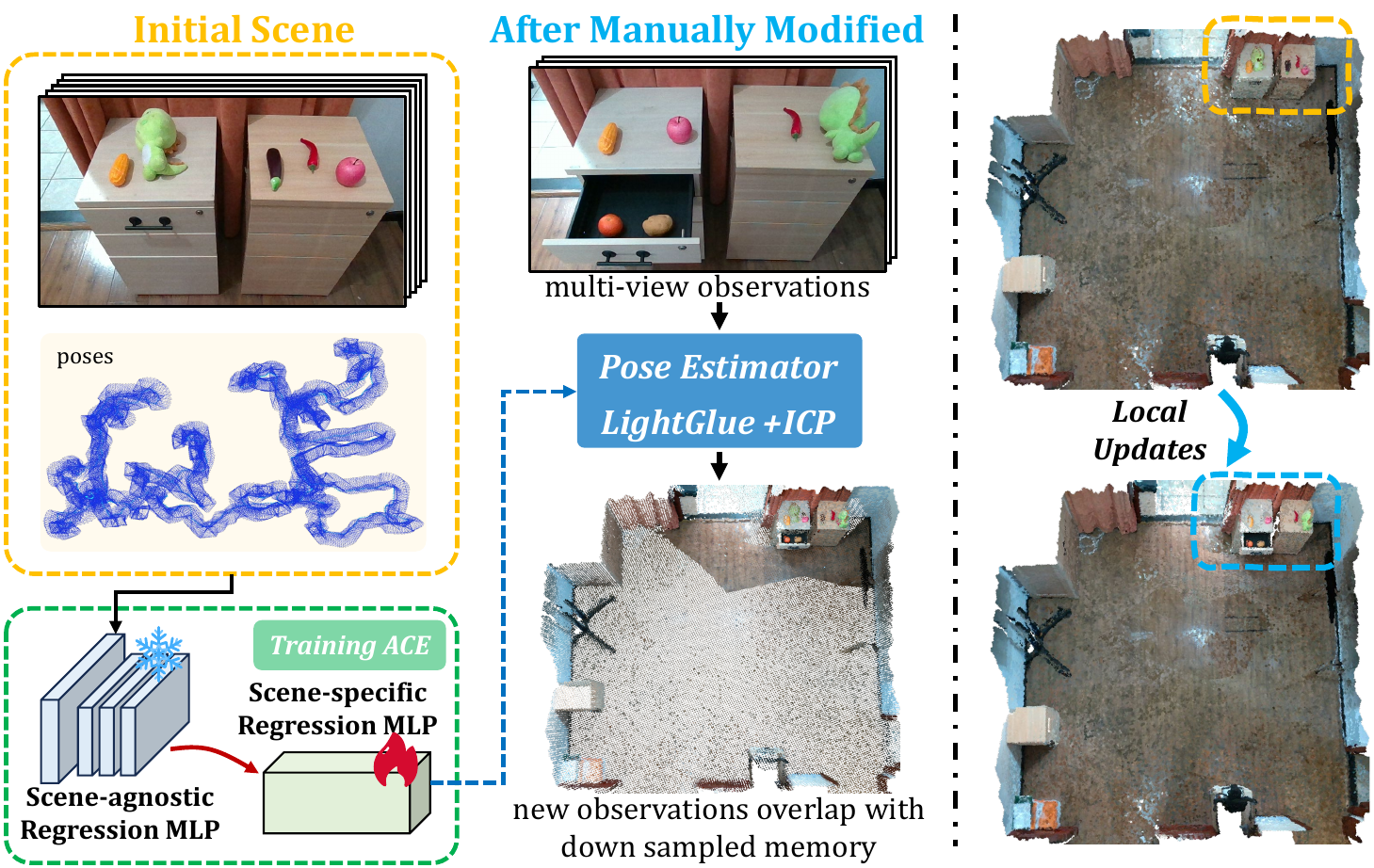}
    \vspace{-6mm}
    \caption{\textbf{Adaptation in interactions with manually modified scenes.} (1) We train the scene-specific regression MLP of the ACE model using RGB images and their poses, making the process highly efficient.
    (2) After manual scene modification, multi-view observations allow rough global pose estimation via ACE, refined further using LightGlue and ICP. The new viewpoint’s point cloud closely aligns with the stored pose.
    (3) The bottom image shows accurate local updates to the scene based on observations from the new viewpoint.
    }
    \label{fig:reloc_and_updates}
    \vspace{-4mm}
\end{figure}

Accurate re-localization is crucial for a mobile robot performing manipulations in dynamic environments to maintain an up-to-date scene graph. To ensure precise local updates to the voxelized map, we employ a multi-stage relocalization process that efficiently integrates new observations into the existing scene representation.
We begin by training an ACE scene-specific regression MLP using memory $\langle \irgb_i, \ipos_i \rangle, i \in \{1, ..., t\}$ (see the top left corner in Fig.\ref{fig:reloc_and_updates}). After the robot collects new RGB-D observations ${\img_{k}}$ for $k \in \{t+1, ..., t+n\}$, where each observation $\img_k = \langle \irgb_k, \idep_k, \icb_k \rangle$ includes the RGB image, depth image, and the camera-to-arm-base transformation, we proceed to estimate the robot's pose within the existing map. Then, leveraging the trained ACE model, we predict a rough global pose $\ipos_k$ for each new RGB image $\irgb_k$ by utilizing the prior mapping data $\langle \irgb_i, \ipos_i \rangle$, where $i \in \{1, ..., t\}$. While ACE provides a coarse alignment, its precision may not suffice for detailed scene updates.
To refine these pose estimates, we perform local feature matching using LightGlue \cite{lindenberger2023lightglue}, identifying the historical image $\irgb_{\hat{k}}$, $\hat{k} \in \{1,...,t\}$ with the most feature correspondences to each new observation $\irgb_k$. This step enhances the robustness of the pose estimation by anchoring it to the most similar view in the prior map.
With the matched image pairs, we extract the corresponding point clouds from the RGB-D data and perform multi-scale colored iterative closest point (ICP) alignment. This refinement step minimizes both geometric and photometric discrepancies between the new observations and the existing map, yielding a precise transformation ${\Transform_k^\text{icp}}$ that aligns the new observations within the global coordinate frame.
The refined poses $\ipos_k$ are then updated using the transformation $\ipos_k \leftarrow \Transform^\text{icp} \ipos_k.$

\subsubsection{Remove obsolete indices}
This step is to identify and remove obsolete voxels from the memory map. We propose an efficient method that leverages new RGB-D observations to update the volumetric representation accordingly.
Given the set of stored voxel indices, each associated with 3D positions and color information, we process each new observation $\img_k$ as follows:
(1) We transform the all voxel point cloud (points) $\{\pcd_{\obj_j}\}$, $j \in \{1,...,J\}$ into the current camera coordinate by:
\begin{equation}
\pcd_{\obj_j}^\text{cam} = (\ipos_k)^{-1} \pcd_{\obj_j}
\end{equation}
(2) The transformed points are then projected onto the image plane using the camera's intrinsic parameters to obtain pixel coordinates:
\begin{equation}
\begin{bmatrix} \mathbf{u}_j, \mathbf{v}_j \end{bmatrix} = \Pi(\pcd_{\obj_j}^\text{cam}),
\end{equation}
where $\Pi$ denotes the projection function. We consider only points that project within the image boundaries and are in front of the camera, that is, $z_{j}>0$, $z_{j}$ is the z-axis coordinates of the $\pcd_{\obj_j}^\text{cam}$.
(3) For each valid projected point $i$, we compute the depth and color difference:
\begin{equation}
\Delta z_{i} = \left| \idep_k\begin{bmatrix} \mathbf{u}_{j}^i, \mathbf{v}_{j}^i \end{bmatrix} - z_{j}^i \right|,
\end{equation}
\begin{equation}
\Delta c_{i} = \left| \irgb_k\begin{bmatrix} \mathbf{u}_{j}^i, \mathbf{v}_{j}^i \end{bmatrix} - \mathbf{c}_{j}^i \right|, 
\end{equation}
where
$\mathbf{c}_j$ is the stored color of $\pcd_{\obj_j}$.
(4) We define thresholds $\delta_z$, $\delta_z'$, and $\delta_c$ for depth and color differences. A point $i$ is marked for deletion if: 
\begin{equation}
\text{point } i \text{ is deleted if } 
\begin{cases}
   \Delta z_i > \delta_z, \\
   \Delta z_i > \delta_z' \ \text{and} \ \Delta c_i > \delta_c.
\end{cases}
\end{equation}

\subsubsection{Update low-level memory}
After the above step, the local scene in the historical low-level memory will be updated to the latest state based on the visual information provided by the new RGB-D observations ${\img_{k}}, k \in {t+1, \dots, t+n}$. We process $\img_{k}$ sequentially in the same way as described in Sec.~\ref{3d_mapping}, and fuse it with the historical $\objset_t$ to $\objset_{t+1}$. The set $\langle \irgb_k, \idep_k, \ipos_k \rangle, k \in {t+1, \dots, t+n}$ will also be updated into $\imgsequence_{t+1}$, providing new reference viewpoints for the next re-localization to ensure long-term localization accuracy.

\subsubsection{Update high-Level memory}
For updating the scene graph, we adopt a simple local sub-graph-based update strategy to avoid redundant global updates. Specifically, (1) We compare the historical object set $\objset_t$ with the updated object set $\objset_{t+1}$ to identify the objects that have been deleted or whose points have been changed, forming the set $\objset_{\text{affected}}$. (2) Next, we examine the edges in $\edgeset_t$ to find the parent objects of all objects in $\objset_{\text{affected}}$ and the child objects of those parent objects. These related objects are also added to $\objset_{\text{affected}}$, and all edges associated with $\objset_{\text{affected}}$ are removed from $\edgeset_t$.
(3) We then identify the objects in $\objset_{\text{affected}}$ that still exist in $\objset_{t+1}$ and locate any newly added objects in $\objset_{t+1}$. Both these types of objects are added to the set $\objset_{\text{need\_process}}$ for further processing.
(4) For each object in $\objset_{\text{need\_process}}$, we recompute the spatial relationships as described in Sec.~\ref{3d_sgg}. This involves determining the relevant edges $\edg$ for each object, updating $\edgeset_{t+1}$ with the new connections and builds $\scenegraph_{t+n}$.

\subsection{Language-guided Task planning}
Benefiting from the system's long-term workability in indoor environments, we integrated the advanced large language model GPT-4o. Based on class-agnostic and highly extensible prompt text descriptions, we decompose long-term tasks described in natural language into multiple subtasks that the robot can easily adapt to. Each subtask output by GPT consists of an ``action\_name'' and multiple ``object\_name'', which are directly extracted from the description and maintain the same level of abstraction as the described objects, ensuring that the original meaning is not distorted during task decomposition.

\subsection{Navigation}
\subsubsection{Localization}
Before initiating each navigation task, we first determine the robot's precise pose as described in Sec.~\ref{reloclizetion_and_refinement}, establishing its position in the world coordinate system (start point). To locate the target object(s), we utilize CLIP to obtain embeddings of the specified object names. If only object A is specified, we compute its CLIP embedding and compare it with the embeddings of all objects in the scene using cosine similarity. The object with the highest similarity score is identified as the target location of A (as illustrated in the upper section of Fig.~\ref{fig:scenegraphinitial}). For tasks involving a spatial relationship between two objects (e.g., ``object A is on object B''), we compute CLIP embeddings for both A and B. We then compare these embeddings with those of the scene objects to obtain similarity scores. For each object, we select the top-$k$ most similar scene objects (top-$k$ A and top-$k$ B, respectively).
Next, we calculate the Euclidean distances between each pair of candidate locations from top-$k$ A and top-$k$ B. The pair with the shortest distance is deemed the most probable configuration, and the location of the corresponding A candidate is selected as the target point.

\subsubsection{Mobile control}
Once the target location is determined, we use the A* \cite{hart1968formal} algorithm to generate a collision-free navigation path from the start point to the target point. The robot then follows this path using a PID controller to ensure accurate and smooth navigation.

\begin{figure}
    \centering
    \includegraphics[width=1\linewidth]{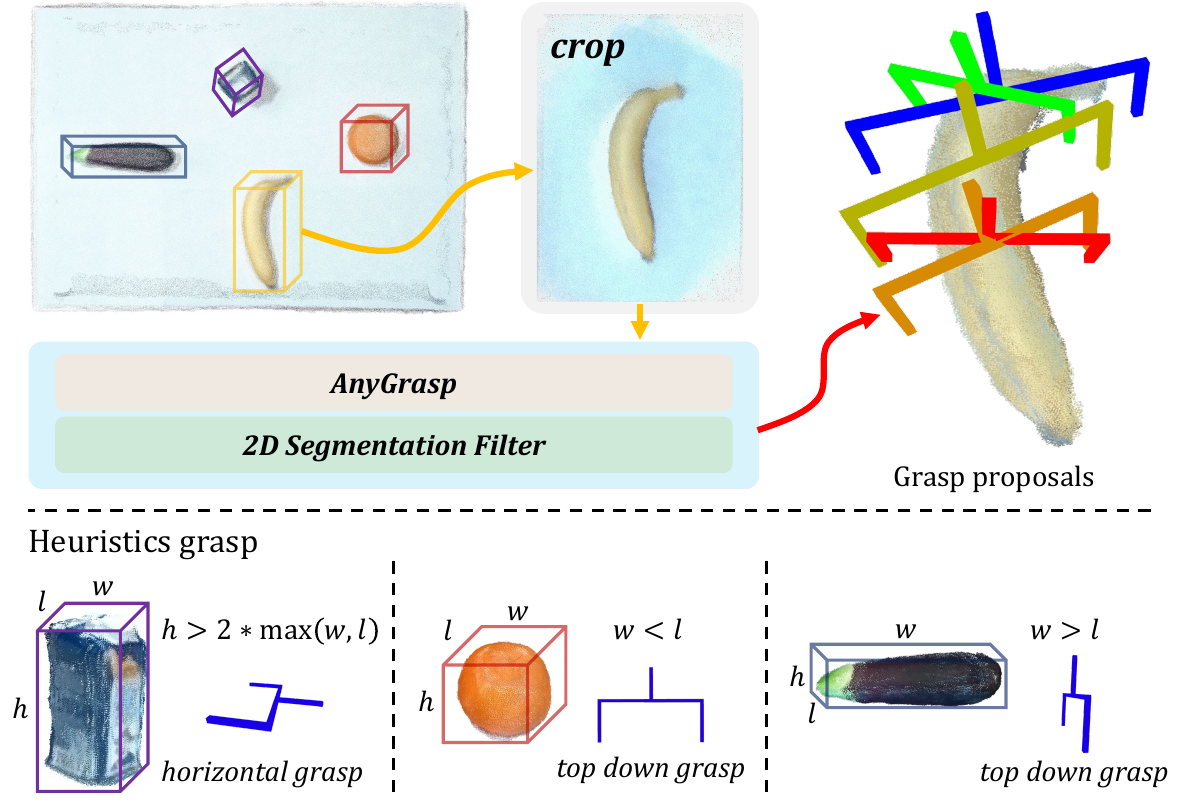}
    \vspace{-6mm}
    \caption{\textbf{Two proposed grasp strategies in \ourname{}}. In the first row, we cropped the point cloud input into anyGrasp within a certain range around the target object, allowing anyGrasp to focus more on the target object without compromising the generation of collision-free grasps. Furthermore, we filtered the grasps based on translational and rotational costs, with the red grasps indicating the highest confidence. In the second row, we show our heuristic grasp strategy, which leverages the object's bounding box information to rotate and select the most appropriate grasp orientation.
    }
    \label{fig:p_strategy}
    \vspace{-4mm}
\end{figure}

\subsection{Manipulation}

Once the robot reaches the target location, we employ a pick-and-place strategy similar to Ok-Robot \cite{Liu_2024}. First, the robot retrieves the object's 3D coordinates from semantic memory and points its camera at the target to capture RGBD images. Then, the robot performs either a ``Pick up'' or ``Place''.

\subsubsection{Pick up}
To focus the AnyGrasp model on the target object, we first preprocess the point cloud by cropping it to a region around the target object, based on its detected bounding box. This step refines the input for AnyGrasp, leading to more accurate and efficient grasp predictions. After generating candidate grasps, we apply cost-based filtering to select the best option. The robot executes the grasp with the highest confidence, leveraging additional segmentation from Grounding DINO and Segment Anything-2 for precise targeting.
Furthermore, we introduce a heuristic grasp strategy that uses the object's bounding box information to rotate and align the gripper for the most suitable grasp orientation, as shown in Fig.~\ref{fig:p_strategy}. This heuristic strategy is only activated when AnyGrasp can't provide a suitable grasp, ensuring optimal interaction with the object’s geometry

\subsubsection{Place}
We first obtain the point cloud $\pcd_a$ of the target object using Grounding DINO and Segment Anything-2 (SAM2), as described in Sec.~\ref{ov2dseg}. This point cloud is then transformed into the robot's base coordinate frame. Next, we compute the median coordinates $x_m$ and $y_m$, and determine the drop height as follows:
\begin{equation} z_{\text{max}} = 0.1 + \max \left( z \mid 0 \leq x \leq x_m, |y - y_m| < 0.1 \right), \end{equation}
where $(x, y, z) \in \pcd_a$. A buffer of 0.1 is added to account for potential collisions. The robot executes the placement manipulation at the computed coordinates $(x_m, y_m, z_{\text{max}})$.

%% file: e_experiments.tex
\section{Experiments}
In this section, We evaluate \ourname{}'s performance in dynamic, real-world environments to answer two key questions: (1) How well does our system adapt to changes by updating the dynamic scene graph? (2) How effectively does this facilitate the completion of consecutive tasks without manual resets?

\begin{figure}
    \centering
    \includegraphics[width=1\linewidth]{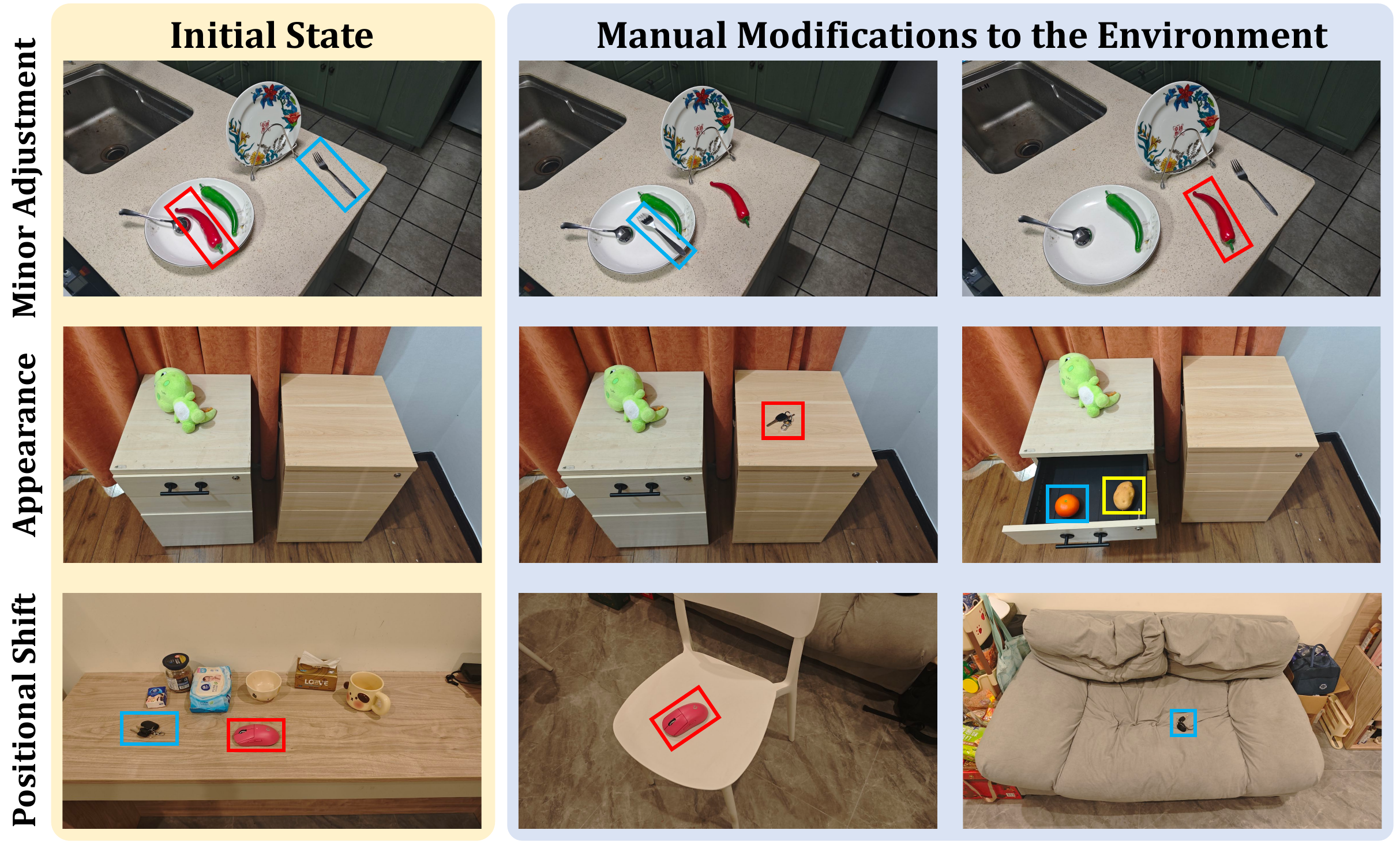}
    \vspace{-6mm}
    \caption{\textbf{Degrees of environmental modifications.} The left column shows the initial state of the scene, while the two columns on the right represent the state of the scene after manual modifications.
    }
    \label{fig:modified_environment}
    \vspace{-4mm}
\end{figure}

\subsubsection{Robot Setups}
We used a real-world setup with a UFACTORY xARM6 robotic arm on an Agilex Ranger Mini 3 mobile base, equipped with a RealSense D455 camera for perception and a basket for item transport.

\subsubsection{Environment and Task Setups}
To verify our method's ability to enable robots to perform long-term tasks in dynamic environments, we designed an experiment in 4 real-world rooms. The experiment simulated dynamic environments through human interactions, which modified object positions, added new objects, or revealed hidden ones. Each experiment began with a 3D scene graph and involved two consecutive tasks. 
Before the first task, we manually adjusted objects related to the second task by changing their positions, adding new objects, or revealing hidden ones. These modifications were designed to be detectable during the execution of the first task. After completing the first task, instead of being repositioned to an initial start point, the robot continued to the next task from its current state. This setup simulated real-world scenarios with continuous environmental changes caused by human interference.

To evaluate the robot's ability to detect and adapt to environmental changes, we categorized the modifications to the objects in the second task into three levels (see Fig.~\ref{fig:modified_environment}):

\textbf{(1) Minor Adjustment:}
Slight position changes that remain detectable from the original location. For instance, a slightly moved object may still be visible within the robot’s field of view.
\textbf{(2) Appearance:}
Previously hidden objects become visible, such as items revealed by opening a drawer or placing a key on the table. These changes introduce new nodes into the scene graph.
\textbf{(3) Positional Shift:}
Significant relocations of objects render them undetectable from their original positions, altering their spatial relationships with other objects in the scene graph.

Overall, for each method, each level of modification was tested through 20 long-term tasks per room across 4 rooms, resulting in a total of 80 trials per modification type. With three modification levels, each method was subjected to 240 long-term task experiments, in which objects or their positions were randomized in every trial.

\begin{table}[h]
\caption{\textbf{Quantitative Results on Dynamic Scene Adaptation and Scene Graph Generation.}}
\vspace{-2mm}
\centering
\resizebox{1\linewidth}{!}{

\begin{tabular}{lcccccc}
\toprule
Task    & \multicolumn{2}{c}{Minor Adjustment} & \multicolumn{2}{c}{Appearance} & \multicolumn{2}{c}{Positional Shift} \\  \cmidrule(lr){2-3} \cmidrule(lr){4-5} \cmidrule(l){6-7}
Metrics & GPT-4o & Ours & GPT-4o & Ours & GPT-4o & Ours \\ \midrule
SCDA & 41.44\% & \textbf{95.37\%} & 64.25\% & \textbf{93.22\%} & 66.35\% & \textbf{94.23\%} \\
SGA & 54.60\% & \textbf{88.75\%} & 52.25\% & \textbf{84.86\%} & 46.18\% & \textbf{83.72\%} \\ \bottomrule
\end{tabular}
}
\label{tab:ssg_results}
\vspace{-5mm}
\end{table}

\subsubsection{Baselines}
To demonstrate our approach's adaptability to changing environments and the effectiveness of scene graph prediction, we evaluated scene changes and scene graph predictions. Using GPT-4o as the base model, enhanced with chain-of-thought (CoT) reasoning \cite{wei2023chainofthoughtpromptingelicitsreasoning}, similar to the method proposed by Jiang et al. \cite{jiang2024roboexpactionconditionedscenegraph}, the model takes RGB observations from the robot, matches them to the most similar image in memory, detects differences, and predicts the scene graph based on the new RGB observation. 
For long-term task execution, we implemented Ok-Robot \cite{Liu_2024} on our mobile robot as a comparison method. 
Additionally, we compared our approach with Ok-Robot and ConceptGraphs \cite{gu2023conceptgraphsopenvocabulary3dscene} in terms of memory usage and scene update time.

\begin{table}[h]
\vspace{-4mm}
\caption{\textbf{Success Rate of Long-term Tasks and Subtasks}}
\vspace{-2mm}
\centering
\resizebox{\linewidth}{!}{
\begin{tabular}{cccccc}
\toprule
Task & Method & Minor Adjustment & Appearance & Positional Shift & \textbf{Total(\%)} \\ \midrule
\multirow{2}{*}{Pick up} & Ok-Robot & 84 / 110  & 61 / 92 & 59 / 87 & 70.58\% \\
 & Ours & 111 / 137 & 111 / 136 & 108 / 133 & \textbf{81.28\%} \\ \midrule
\multirow{2}{*}{Place} & Ok-Robot & 53 / 64 & 40 / 51 & 42 / 51 & 81.32\% \\
 & Ours & 80 / 93 & 83 / 95 & 74 / 85 & \textbf{86.81\%} \\ \midrule
\multirow{2}{*}{Navigation} & Ok-Robot & 179 / 210 & 146 / 184 & 137 / 180 & 80.48\% \\
 & Ours & 228 / 236 & 239 / 254 & 224 / 245 & \textbf{94.01\%} \\ \midrule
\multirow{2}{*}{\textbf{Long-term}} & Ok-Robot & 12 / 80 & 0 / 80 & 0 / 80 & 5.00\% \\
 & Ours & 33 / 80 & 28 / 80 & 23 / 80 & \textbf{35.00\%} \\ \bottomrule
\end{tabular}
}
\label{tab:sr_results}
\vspace{-1mm}
\end{table}

\subsubsection{Evaluation}
To thoroughly evaluate \ourname{}'s adaptability to dynamic environments and its effectiveness in long-term tasks, we involved human evaluators in modifying the environment, constructing ground truth (GT) scene graphs, and determining task execution success. Our evaluation focuses on three key aspects:
\textit{\textbf{Dynamic Scene Adaptation and Scene Graph Generation:}} We recorded RGB observations during the robot's task execution and compared \ourname{}'s performance with baseline methods in detecting scene changes and generating scene graphs. \textit{\textbf{Long-Term Tasks Evaluation:}} In addition to assessing the overall success rate of long-term tasks, we evaluated the success rates of individual subtasks such as picking, placing, and navigation. This detailed evaluation provides a comprehensive analysis of the effectiveness of our method across different components of task execution. The three main metrics used for evaluation are as follows:
\textbf{(1) Scene Change Detection Accuracy (SCDA):} This metric measures the accuracy of detecting changes in the observed scene and matching them to the corresponding memory-stored RGB information.
\textbf{(2) Scene Graph Accuracy (SGA):} 
This metric evaluates whether the generated scene graph matches the GT scene graph and only recognizes three types of relations: ``on'', ``belong'', and ``inside''.
If the final scene graph perfectly matches the GT version, the experiment is considered successful and assigned a score of 1, otherwise 0.
\textbf{(3) Task Success Rate:} This metric represents the overall task completion success rate. For long-term tasks, success is only counted if all subtasks are completed successfully.

\subsubsection{Results}
As shown in Tab.~\ref{tab:ssg_results}, in the SCDA evaluation, GPT-4o often struggles with inconsistencies between the scope of historical observations and new observations, which introduces a significant amount of noise in the responses, 
leading to poor performance.
This inconsistency also makes accurate detection under ``Minor Adjustment'' scenarios very challenging. In contrast, \ourname{}, supported by precise re-localization, can accurately identify the voxel index where changes have occurred in the scene, significantly outperforming the baseline.
Specifically, in ``Minor Adjustment'' scenarios, \ourname{} exceeds the baseline by nearly 54\%, primarily because GPT-4o struggles to recognize smaller positional changes. Additionally, in ``Appearance'' and ``Positional Shift'' scenarios, \ourname{} achieves a scene change recognition success rate approximately 28\% higher than the GPT-4o.
In the SGA evaluation, both approaches generate scene graphs based on new observations, with SGA accuracy in ``Minor Adjustment'' scenarios being higher than in ``Appearance'' and ``Position Shift'' scenarios. The main issue with our method arises when the same object is decomposed into multiple nodes, affecting the scene graph's accuracy. On the other hand, GPT-4o struggles to accurately recognize relationships or redundantly defines objects, leading to erroneous results.

In Tab.~\ref{tab:sr_results}, we have the fellow observations:
\textbf{(1)} In the ``Minor Adjustment'' environment, although the object is slightly moved, it remains within the robot’s field of view. This makes it highly likely for the robot to navigate near the target, resulting in a significantly higher success rate compared to ``Appearance'' and ``Positional Shift'' (in 80 trials, it achieved 5 and 10 more successes, respectively). In the ``Positional Shift'' scenario, the residual effect of CLIP features can occasionally mislead the robot into navigating toward the object’s historical location, ultimately causing navigation failure. In contrast, for ``Appearance'', where a new object emerges, the robot does not face the challenge of misjudging the original object’s position, generally leading to a higher success rate than ``Positional Shift'' (with 5 more successes out of 80 trials).
\textbf{(2)} \ourname{} demonstrates significantly enhanced manipulation capabilities over Ok-Robot. For the "Pick up" task, compared to Ok-Robot, \ourname{} employs two proposed grasp strategies that focus specifically on the target object. By optimizing the selection of grasp candidates and integrating a heuristic grasping method, \ourname{} reduces environmental interference and ensures the robot selects the optimal grasp. This results in a 10.7\% higher pick-up success rate than Ok-Robot, which relies solely on AnyGrasp. Regarding ``Place'' task, \ourname{} utilizes a lower placement position and an inclined placement method (see our \href{https://bjhyzj.github.io/dovsg-web}{project page} for details), achieving an overall success rate that is 5.49\% higher than Ok-Robot.
\textbf{(3)} In dynamic environments, \ourname{} significantly outperforms Ok-Robot (which assumes a static scene) in long-term tasks, thanks to its ability to adapt to scene changes. Although Ok-Robot can occasionally succeed in locating the correct object under minor changes (e.g., ``Minor Adjustment''), it struggles with larger modifications such as ``Appearance'' or ``Positional Shift'' because it cannot partially update its scene representation—making success in these scenarios nearly impossible. As a result, Ok-Robot’s success rate for long-term tasks in dynamic environments is approximately 30\% lower than \ourname{}.

Tab.~\ref{tab:eff_comp} further demonstrates the memory consumption required to store memory at a 1 cm resolution and the update times for \ourname{}, Ok-Robot, and ConceptGraphs after processing 1200 frames of RGBD data within a $40\,\mathrm{m}^2$ scene. It is important to note that since Ok-Robot and ConceptGraphs cannot perform partial updates, the table only presents the time required for a complete scene update. The results show that Ok-Robot consumes approximately 13 times more memory than \ourname{} and has update times 20 times longer. While ConceptGraphs and \ourname{} have similar memory usage due to both using 3D scene graphs, \ourname{}'s ability to perform local updates allows it to achieve update times 27 times faster than ConceptGraphs.

\begin{table}
\caption{\textbf{Effieciency comparison.}}
\vspace{-2mm}
\centering
\resizebox{0.7\linewidth}{!}{
\begin{tabular}{lcc}
\toprule
Method ($\text{m}^2$) & Memory (GB)$\downarrow$ & Time (min)$\downarrow$ \\ \midrule
Ok-Robot & 2 & 20 \\
ConceptGraphs & \textbf{0.15} & 27 \\
Ours & \textbf{0.15} & \textbf{1} \\ 
\bottomrule
\end{tabular}
}
\label{tab:eff_comp}
\vspace{-5mm}
\end{table}

%% file: f_limitations.tex
\section{Limitation and Future Work} 
\textit{Limitation:} The performance of \ourname{}'s visual relocalization depends on the presence of distinctive visual cues (e.g., textures, edges, objects) for key point matching. In environments with sparse, minimal, or repetitive features, the system may struggle to extract enough distinctive points, impacting localization accuracy.

\textit{Future work:} Exploring real-time precise localization through multi-sensor fusion, efficient memory update mechanisms, and developing lightweight methods for scene representation is crucial. Additionally, enabling efficient collaboration between mobile robots and manipulators will be key to making mobile manipulation more practical in real-world scenarios.

%% file: g_conclusions.tex
\section{Conclusion}
In this paper, we introduce \ourname{}, an innovative framework designed to enable mobile robots to perform long-term tasks in dynamic environments by continuously local updates of 3D scene graphs. \ourname{} enables the robot to adapt to changes caused by human interaction or by the robot's own task execution. These capabilities ensure the precise execution of long-term tasks without being affected by cumulative temporal errors. By evaluating long-term tasks and subtasks such as pickup, place, and navigation, our results highlight the robustness and effectiveness of \ourname{} in handling the complexity of dynamic real-world scenes. 
Compared to methods that assume static environments and cannot adjust to scene changes, \ourname{} achieves a 30\% higher success rate in long-term tasks, accelerates updates by up to 27 times in typical room scenarios, and reduces memory consumption by a factor of 13.